# DUET: Decoupled Upper- and Lower-Body Execution for Task-Agnostic Humanoid Loco-Manipulation

Kapalik Khanal, Biswash Khatiwada, Stephen Afrifa, Fan Wu, and Ramesh Bahadur Bist*
*Department of Biological and Agricultural Engineering, North Carolina State University, Raleigh, North Carolina, USA*
*Corresponding author: rbbist@ncsu.edu

**Abstract—** Changing a humanoid robot's manipulation policy should not require retraining the controller that maintains the robot's balance. This study introduces DUET, a decoupled execution framework that enables locomotion and manipulation controllers to operate simultaneously by combining their commands only at the robot interface. A reinforcement-learning locomotion policy controls the 12 leg joints and a waist yaw joint, tracking planar velocity, turning rate, and pelvis height from deep crouching to upright standing on a 23-degree-of-freedom Unitree G1. The 10 arm joints are excluded from the action space, while their positions and velocities are included only as observations, so that any compatible manipulation policy that generates arm-joint targets can control the upper body, and the locomotion policy keeps the lower body stable without requiring shared training or direct coordination. This study demonstrates the modular approach on hardware by pairing the same locomotion checkpoint with two different vision-language-action models, GR00T N1.7 and pi0.5. In simulation evaluation, the locomotion controller kept the robot's lower body stable with no falls while tracking the commanded planar velocity, yaw rate, and pelvis height with errors of 0.186 m/s, 0.142 rad/s, and 0.016 m, respectively. Ablation studies identify the upper body pose curriculum as the dominant component, and removing this training step reduces episode survival from 20.0 to 7.4 seconds and increases the fall rate to 6.06 per environment minute. Robustness tests show no falls under pushes of up to 1.0 m/s within the trained payload range, and no falls at twice the payload limit for pushes below 0.7 m/s. These results show that a manipulation policy can be exchanged without retraining, restarting, or otherwise touching the layer that keeps the robot upright.

## I. INTRODUCTION

Humanoid robots are increasingly expected to perform manipulation while remaining mobile. Yet locomotion and manipulation are often developed under different objectives and later integrated into a single robotic system [1]. Locomotion controllers typically optimize for dynamic stability under prescribed body motions, whereas manipulation policies are trained to generate task-specific upper-body behaviors [1], [2]. Combining the two control components often requires joint training, whole-body policy design, or explicit coordination between controllers, making the resulting system difficult to modify when the manipulation task or policy changes. Although such approaches can produce coordinated behavior, they can also create dependencies between the manipulation policy and the controller responsible for maintaining balance. As a result, changing the manipulation behavior may require modification, adaptation, or retraining elsewhere in the control stack.

A more modular alternative is to treat lower-body stabilization as a reusable capability that can remain unchanged while independently developed manipulation policies control the upper body.

The challenge of maintaining such modularity is particularly evident in tasks that require a robot to move between work locations and manipulate objects at different heights. For example, a humanoid transporting agricultural produce between workstations or retrieving fruit from a low table requires the robot to approach the workspace, adapt its body configuration to the target, perceive the object, and execute a grasp, the same sequence that mobile manipulation systems on legged [3] and wheeled [4] platforms must coordinate. Wheeled manipulators address mobility effectively on relatively structured and accessible surfaces, while a substantial portion of humanoid manipulation research evaluates manipulation with the robot fixed or operating from a largely stationary base [5]. Such assumptions become restrictive in human-centered environments, where objects may be located near the floor, on workbenches, or on elevated shelves [6]. Humanoid morphology is well suited to these settings because both the robot and the workspace share similar spatial scales and access requirements. At the same time, advances in learning-based control have substantially improved the capabilities required on both sides of this problem. Massively parallel simulation [7], [8] turned proximal policy optimization [9] into a reliable recipe for legged locomotion, building on earlier reinforcement-learning results on quadrupeds [10], [11], and the recipe was later carried to humanoids trained for the real world [12]. Manipulation has progressed in parallel through imitation learning and more recently vision-language-action (VLA) models capable of transferring learned manipulation behaviors to physical robotic systems [5], [13], [14].

Existing humanoid loco-manipulation systems broadly address this composition in two ways. The first trains a unified, whole-body policy across most or all robot joints, often using human motion or whole-body demonstrations [15], [16], [17], [18]. Such approaches can produce expressive and highly coordinated behavior because upper- and lower-body motion are optimized within a common policy representation. The second keeps legs and arms on separate controllers. The legs run a reinforcement learning (RL) policy trained to maintain balance, while a separate policy drives the arms. HOMIE drives them from an exoskeleton teleoperation cockpit [2], and the decoupled controller released for GR00T N1.5 and N1.6 drives them by inverse kinematics [19]. This study utilizes the second route and focuses on the part that matters for putting a foundation model on legs. Whatever drives the arms should be free to be any learned policy, and exchanging it should cost nothing on the locomotion side. This motivates a framework in which any compatible learned policy can control the arms without requiring retraining or modification of the locomotion controller.

This study introduces Decoupled Upper- and Lower-Body Execution for Manipulation-Agnostic Humanoid Loco-Manipulation (DUET), a framework designed around this requirement. DUET runs an RL locomotion policy and a VLA manipulation policy simultaneously on a 23-DOF Unitree G1, sharing nothing but the robot's body. The locomotion policy owns the 12 leg joints and the waist yaw. It tracks velocity, turning rate, and a commanded pelvis height that ranges from a normal stand to a deep crouch, and it never sees the arms as actions, only as observed joint states. The coupling between the two policies amounts to a few lines in the low-level controller. The locomotion process owns the robot's command topic and copies the arm section of its outgoing message from whatever the manipulation process last published on a separate topic. Since that message is their only point of contact, the locomotion policy remains independent of the arm policy being used, allowing two separately trained VLAs to run with the same locomotion checkpoint without modification.

The key contributions of the proposed framework are to:

1. Develop a velocity- and height-conditioned locomotion policy for the 23-DoF Unitree G1 humanoid.

2. Enable the locomotion layer to remain agnostic about the policy that drives the arms.
3. Isolate the contribution of each of the three training components through matched-budget ablation studies.
4. Evaluate robustness under varying payloads and push disturbances and quantify the postural cost of decoupling through a matched comparison with whole-body control.

The central contribution of DUET is therefore not simply that the arms and legs are controlled separately. DUET makes the identity of the manipulation policy irrelevant to the locomotion policy. Because the locomotion policy learns stabilization over the physical envelope created by manipulation rather than coordinating with one specific manipulation controller, the lower body becomes a reusable mobility layer beneath independently developed manipulation models. This enables improvements in manipulation to be incorporated without retraining the controller responsible for keeping the humanoid upright.

## II. RELATED WORK

### *A. Learned legged locomotion*

Policy-gradient methods trained in massively parallel simulation have become a standard approach for training legged locomotion policies. Isaac Gym [8] and the training recipe of Rudin et al. [7] made thousands of parallel environments practical on a single GPU, following earlier reinforcement-learning controllers for quadrupeds [10], [11], and the same recipe was later carried to humanoids operating outdoors [12]. Sim-to-real transfer relies on domain randomization [20] and, in some systems, on explicit adaptation to unobserved dynamics [21]. The locomotion policy presented here follows this line but is trained in a MuJoCo-based pipeline (mjlab) [22], [23], rather than Isaac Gym, using the GPU-accelerated batched simulation that MuJoCo Playground [24] made practical to reach a comparable environment count on a single GPU.

### *B. Whole-body humanoid control*

A second line of work learns coordinated upper- and lower-body behavior within a unified whole-body policy, often using retargeted human motion or whole-body demonstrations. Expressive Whole-Body Control [15], HumanPlus [16], OmniH2O [17], and HOVER [18] produce coordinated behavior across arms and legs and support teleoperated or retargeted commands. This integration is powerful because balance and manipulation are learned within a shared control representation, allowing the entire body to contribute to task execution. However, the same coupling limits modularity: upper-body behavior is embedded within the learned whole-body controller, making it difficult to replace the manipulation policy independently without adapting or retraining the controller. DUET examines the opposite design choice, sacrificing some degree of whole-body coordination to make manipulation-policy exchange possible while keeping the locomotion policy fixed.

### *C. Decoupled loco-manipulation*

Closest to the present setting are systems that separate the two controllers. HOMIE [2] trains a lower-body policy that tracks velocity and height while the upper body is driven by an isomorphic exoskeleton cockpit, and it introduces an upper-body pose curriculum, a height-tracking reward, and mirror-symmetry augmentation in training. The decoupled controller released for GR00T N1.5 and N1.6 [19] drives the arms through inverse kinematics. This study adopts the three training components from HOMIE, reproduces them in a different simulator on a different robot variant, and changes the source of the arm commands: instead of a cockpit or an IK solver, the arm targets come from a pretrained VLA that has no knowledge of the lower body.

*D. Vision-language-action models*

Large multimodal policies trained on robot demonstrations now transfer to new manipulation tasks with limited data. RT-2 [25], OpenVLA [14], the π family [13], [26], and GR00T N1 [5] share a common interface: they take images and a language instruction and emit robot actions, either as discretized action tokens [14], [25] or as continuous action chunks from flow-matching or diffusion-based action heads [5], [13], [26] that build on architectures developed for visuomotor control [27], [28]. These models are trained primarily for arm manipulation and are most often evaluated on a fixed or stationary base, so the joint targets they emit assume a stationary frame. The present study treats such a model as an opaque publisher of those targets and asks what the legs must be trained for so that it can run unchanged on a walking robot.

*E. Positioning*

**Table I** summarizes how the three threads compare. They differ in where the arms sit relative to the learned controller and in what becomes of the arm targets once they arrive. Whole-body systems place every joint within a single learned action space. Some expose an interface an autonomous policy can drive, such as HOVER's joint-angle mode or the end-effector trajectories of UMI on Legs. In both cases, the controller still emits all joint commands itself, so an incoming target is a tracking objective it may deviate from when balance conflicts. Decoupled systems keep the arms outside the action space and pass targets through unaltered, but the driver has been a human operator or a kinematic solver rather than an interchangeable pretrained model.

**TABLE I:** Comparison with related loco-manipulation systems

| System | Architecture | Arm Driver | Height Command | Arm Command Handling | Swappable Manipulation Policy |
|---|---|---|---|---|---|
| EWBC [15] | Whole body | Retargeted human motion | Yes | Tracked by a learned controller | No |
| HumanPlus [16] | Whole body | RGB shadowing / learned skill policy | No | Tracked by a learned controller | Yes, task-specific skill policies |
| OmniH2O [17] | Whole body | Teleoperation (VR/RGB) | No | Tracked by a learned controller | Yes, keypoint targets |
| HOVER [18] | Whole body | Multi-mode commands | Yes | Tracked by a learned controller | Yes, joint-angle mode |
| HOMIE [2] | Decoupled | Exoskeleton cockpit | Yes | Passed through unchanged | Not demonstrated |
| GR00T Decoupled WBC [19] | Decoupled | Inverse kinematics | Yes | Passed through unchanged | No |
| UMI on Legs [3] | Whole body, task-space interface | Learned policy, EE trajectories | No | Tracked by a learned controller | Yes |
| DUET (Proposed) | Decoupled | Learned VLA, joint targets | Yes | Passed through (rate-limited) | Yes |

DUET is the only entry combining the two. Arm targets reach the joints without being modified by the locomotion policy, and the policy producing them can be exchanged. This matters for a manipulation model trained on a fixed base, which learned its action distribution assuming its targets are executed as issued. The comparison is not a ranking: HOVER spans several control modes, whereas the present system addresses one, and whole-body policies retain the arms as momentum actuators for postural control, a capability DUET gives up at a cost quantified in Section VI-E.

## III. SYSTEM ARCHITECTURE

The robot in this study is a Unitree G1 in its 23-DOF configuration [29], with six joints per leg, one waist yaw joint, and five joints per arm ending at the wrist roll. Both wrists carry a Unitree Dex3-1 three-fingered hand on a 3D-printed bracket, and the right bracket also holds a wrist camera. Vision comes from two RealSense cameras: a D455 on the head for a wide view of the scene and a D405 at the right wrist for close range during grasping. Each hand assembly weighs 810 g, of which the hand accounts for 710 g. The Dex3-1 fingers have their own command topic and are not among the 23 joints considered here.

The upper half of Fig. 1 shows how the locomotion policy is produced. Training runs in parallel simulation against the upper-body pose curriculum of Section IV-C and the randomized dynamics of Section IV-F, with the arms driven by a disturbance generator rather than by any controller the policy can influence. The policy is exported to ONNX with a fixed interface of 71 inputs and 13 outputs, and the joint order, PD gains, action scales, observation layout, commanded ranges, and a hash of all of them are written into the file. The export is therefore self-describing, which matters because an ONNX graph alone records nothing about the gains or joint ordering it assumes.

The lower half of Fig. 1 shows deployment. Two processes run on a workstation connected over Ethernet and communicate by DDS. The locomotion process loads the exported policy and is the only writer of the robot's low-level command topic; the manipulation process publishes arm joint targets to a second topic that reuses the same message type. Each cycle at 50 Hz, the locomotion process reads the state of all 23 joints, assembles the 71-dimensional observation, evaluates the policy for 13 lower-body offsets, writes those into the outgoing message, and fills the ten arm entries from the most recent message on the arm topic together with the gains carried in it, so the manipulation policy retains control of its own compliance. The assembled message is published once and tracked by the onboard PD controllers at 200 Hz. The manipulation policy emits action chunks at roughly 15 Hz, and the locomotion process holds the last received targets between arrivals rather than waiting on them, so a stalled manipulation process cannot stall the legs.

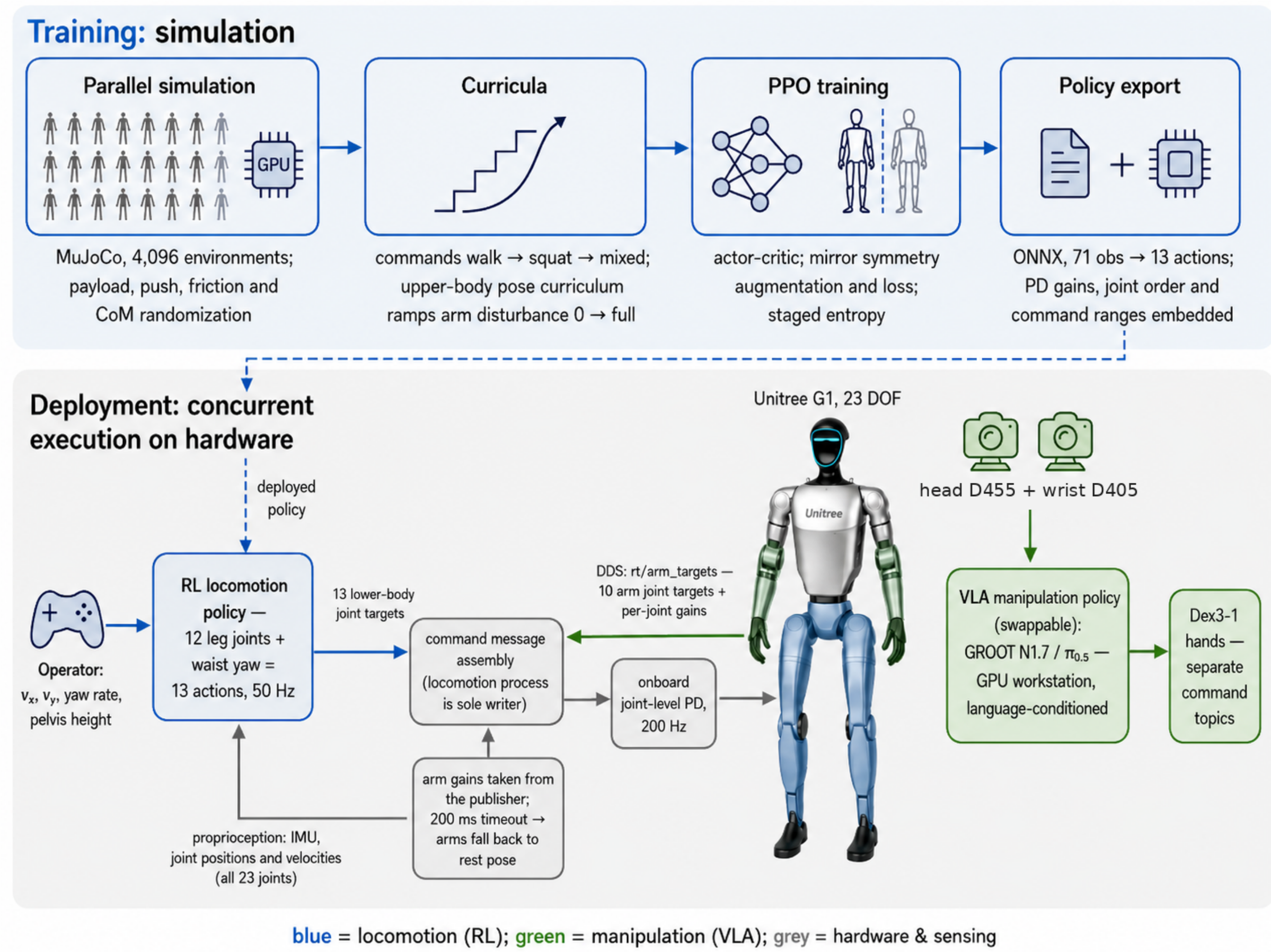

*Fig. 1. System overview. Top: the locomotion policy is trained in parallel simulation against an upper-body pose curriculum and randomized dynamics, then exported to ONNX together with the joint order, gains, action scales, observation layout, and command ranges it was trained against. Bottom: at deployment the locomotion process is the sole writer of the robot's low-level command message. Each cycle it fills the thirteen lower-body entries from the policy and the ten arm entries from the most recent message published by the manipulation process on a separate topic, and the assembled message is tracked by the onboard joint-level PD controllers. The two policies exchange nothing with each other.*

Keeping the arm channel this thin is what makes the manipulation policy interchangeable. The locomotion policy never receives the arm targets or a task description, has no representation of which model is publishing, and observes the arm joints only through their measured positions and velocities, in the same way it observes any other disturbance. Exchanging one manipulation policy for another changes which process writes to the arm topic and nothing else, with no retraining, reconfiguration, or restart of the locomotion process, and no learned component running on the robot itself.

## IV. LOCOMOTION POLICY

### *A. Observations and actions*

The policy acts at 50 Hz on thirteen joints: the twelve leg joints and the waist-yaw joint. Its action is a per-joint position offset added to a fixed nominal pose and scaled per joint, and the resulting targets are tracked by joint-level PD controllers running at 200 Hz. The proportional gains are 40.2 N·m/rad for the hip pitch, hip yaw, and waist joints, 99.1 for the hip roll and knee joints, and 28.5 for the ankles, with damping of 2.6, 6.3, and 1.8 N·m·s/rad, respectively. Action scales are 0.55 rad for the hip pitch, hip yaw, and waist joints, 0.35 for the hip roll and knee joints, and 0.44 for the ankles.

The observation is a 71-dimensional vector containing base angular velocity, projected gravity, the three-dimensional velocity command, a two-dimensional gait phase clock, the positions and velocities of all twenty-three joints, the previous thirteen-dimensional action, and the commanded pelvis height. The arm joints, therefore, appear in the observation as 20 of its entries but never in the action. The critic additionally receives base linear velocity and foot-level quantities that are unavailable on hardware. Both networks are multilayer perceptrons with hidden sizes of 512, 256, and 128 and ELU activations, with running normalization of the inputs.

### *B. Commands*

Three commands are resampled at intervals between three and eight seconds: planar linear velocity, yaw rate, and pelvis height. Pelvis height is measured from the root link to the lower of the two feet so that the reference does not move when the swing foot leaves the ground. The commanded height ranges from 0.18 m to 0.73 m. Deep crouches are commanded only while the robot is stationary; when a locomotion command is active, the commanded height is restricted to lie at or above 0.60 m, which prevents locomotion commands from being issued at deeply crouched configurations. A fraction of environments receive a zero-velocity command so that stationary behavior under arm motion is trained explicitly. Fig. 2 shows the commanded height and the commanded arm pose being held independently on hardware.

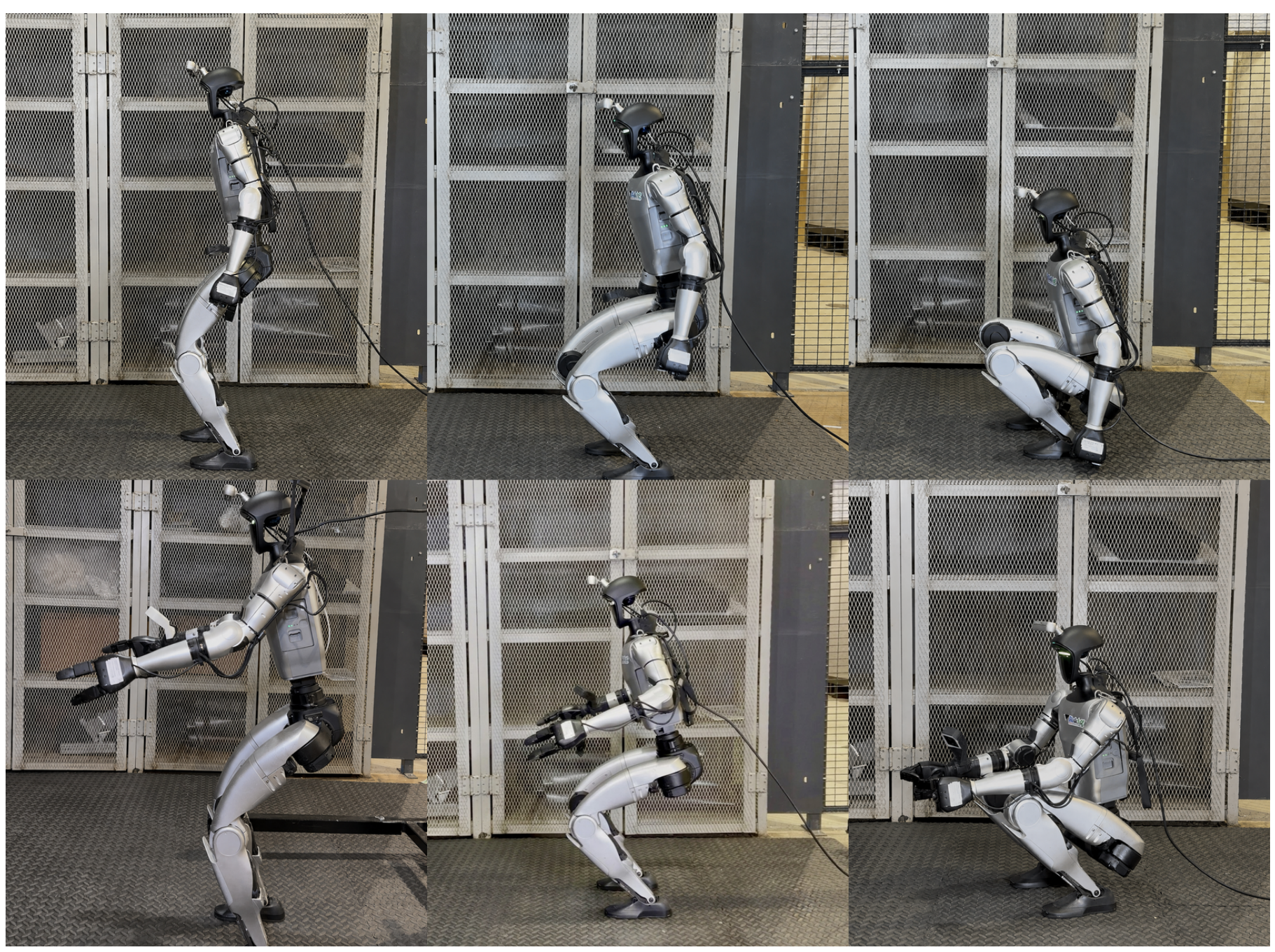

*Fig. 2. Commanded pelvis height and commanded arm pose are independent. Columns are full stand, half crouch, and full crouch. Top row: arms at rest. Bottom row: the same three heights holding an identical commanded forward reach.*

### *C. Upper-body pose curriculum*

During training, a random disturbance generator drives the arms rather than a policy. Every fifty control steps, roughly one second, each environment samples a new arm goal configuration and interpolates towards it. The motion is generated at random and is not replayed from any manipulation trajectory. The anchor fix is the region of configuration space the arms visit while manipulating. Seven goals in ten are sampled around one of six anchors and perturbed by bounded noise: a box held at waist height, a box held at chest height, a forward reach for a bench pick, a low reach paired with a crouch, a high reach, and the resting carry pose. The remaining goals are sampled uniformly from a bounded workspace around the nominal arm configuration so that the policy is not tuned only to the anchors. The amplitude of this motion is ramped from zero to full over the first part of training, and the rate of arm motion is reduced while a locomotion command is active, since violent arm motion during walking does not occur in the intended deployment. Fig. 3 shows the same stride at two commanded heights.

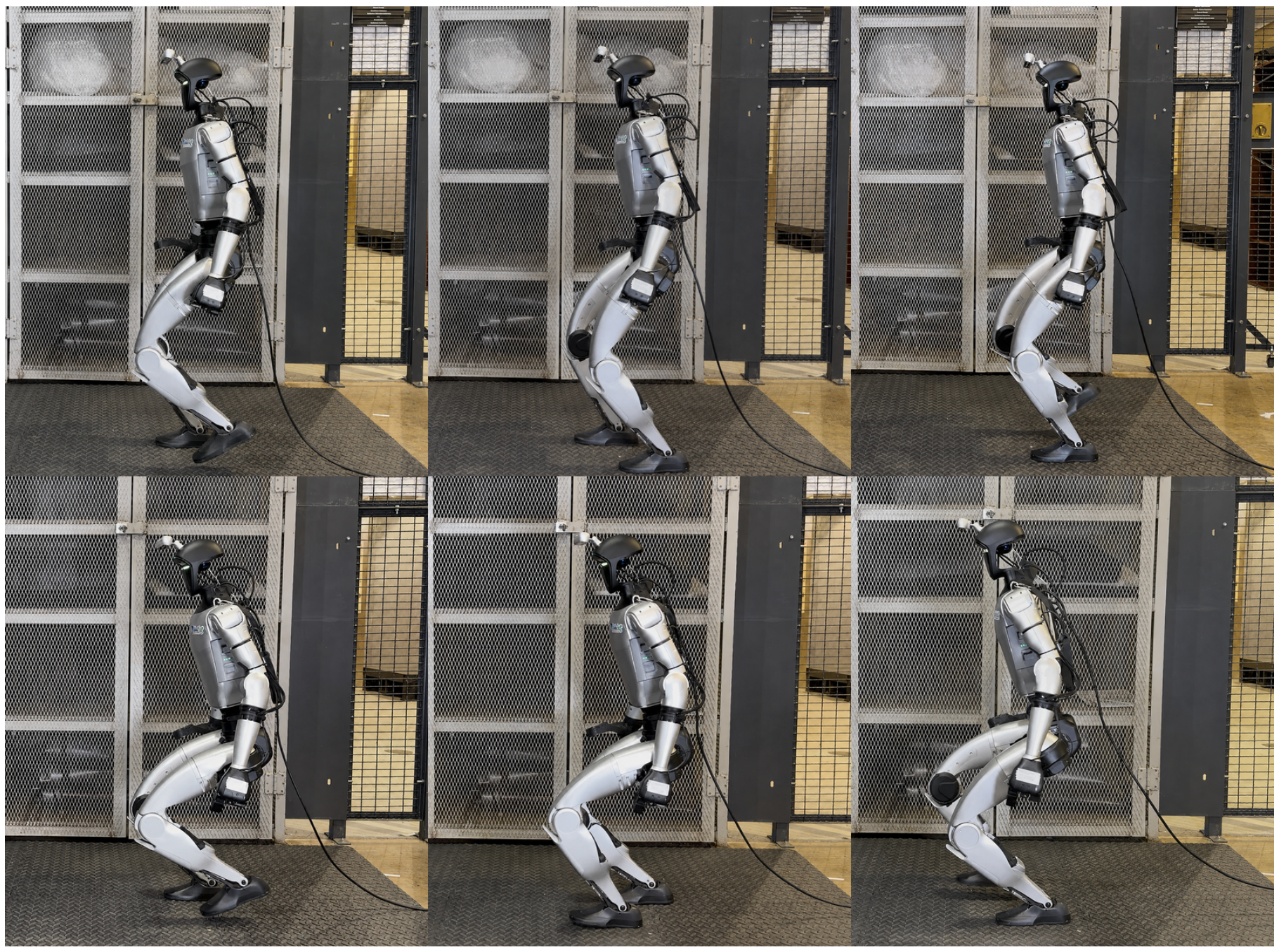

*Fig. 3. Walking at two commanded heights. Top row: a stride at full standing height. Bottom row: the same stride at a commanded pelvis height of 0.60 m. Frames run left to right.*

### D. Rewards

The reward set contains twenty terms, listed in Table II. Three of them define the task: tracking of commanded planar velocity, tracking of commanded yaw rate, and tracking of commanded pelvis height. A second group keeps the robot upright and stable while the arms move, and a third shapes the gait. The remainder are regularizers on joint limits, joint velocity and acceleration, action rate, and self-collision, together with a one-off penalty on termination. Height tracking uses an exponential of the absolute height error measured against the stance foot. The posture regularizer applies looser tolerances to the hip pitch, knee, and ankle pitch joints as lower heights are commanded, so that it does not oppose the fold that a crouch requires.

**TABLE II:** Reward terms and weights.

| Term | Weight | Purpose |
|---|---|---|
| track_linear_velocity | 1.5 | commanded planar velocity |
| track_angular_velocity | 1.75 | commanded yaw rate |
| track_base_height | 2.0 | commanded pelvis height |
| body_orientation | −1.5 | torso upright, relaxed when crouching |
| body_angular_velocity | −0.05 | torso pitch and roll rate |
| posture | 1.0 | speed- and height-scheduled joint deviation |
| feet_distance | −0.75 | stance width band |
| idle_base_motion | −1.0 | base velocity at zero command |
| idle_position_anchor | −5.0 | base displacement at zero command |
| idle_feet_still | −0.25 | foot motion at zero command |
| foot_gait | 0.5 | contact against phase clock |
| foot_clearance | −1.0 | swing height |
| foot_slip | −0.25 | sliding contact |
| feet_air_time | 0.05 | single-stance duration |
| is_terminated | −200 | fall |
| self_collisions | −0.25 | hard self-contact |
| joint_pos_limits | −10.0 | soft limit barrier |
| action_rate | −0.1 | action smoothness |
| joint_vel | −1e−4 | joint velocity |
| joint_acc | −2.5e−7 | joint acceleration |

### *E. Symmetry*

The G1 is nominally symmetric about its sagittal plane, and this is used during training in two ways. Every sampled transition is augmented with its mirror image, obtained by a per-term index permutation and sign flip constructed from the live observation layout and the robot's joint names. In addition, the actor is penalized for disagreement between its output on an observation and the mirror of its output on the mirrored observation. The mirror map is verified at start-up to be an involution on real observations, since a permutation-and-sign map is a valid reflection only if applying it twice is the identity; an incorrect rule would otherwise inject systematically corrupted samples without failing.

### *F. Domain randomization and training*

Randomization covers the quantities that differ most between simulation and the hardware deployment environment and is summarized in Table III. Foot friction, base center-of-mass offset, encoder bias, hand and torso payload, and hand center-of-mass offset are drawn per environment, and planar pushes are applied at intervals. The hand payload is added on top of the simulated hand so that the sampled range brackets the mass of the real hand assembly with and without carried cargo. Training uses proximal policy optimization across 4096 parallel environments with twenty-four steps per environment per update, an adaptive learning rate targeting a fixed KL divergence, and episodes of twenty seconds terminated early if the torso tilts beyond sixty degrees. The entropy coefficient is held constant while the curricula are progressively expanding the task distribution and is then reduced, rather than being fixed for the whole run. All curricula are indexed by the global environment step counter, which is written into each checkpoint, so that a resumed run continues on the distribution it left rather than restarting the ramps.

**TABLE III:** Domain randomization.

| **Quantity** | **Range** |
|---|---|
| Foot friction | 0.3 – 1.6 |
| Base CoM offset (x, y, z) | ±0.10, ±0.06, ±0.05 m |
| Hand payload, per hand | +0.25 – +1.75 kg |
| Hand CoM offset | ±0.05 m, all axes |
| Torso payload | −0.5 – +2.0 kg |
| Encoder bias | ±0.015 rad |
| Planar push, every 3–5 s | ±0.7 m/s |
| Linear velocity command, x | −0.8 – 1.2 m/s |
| Linear velocity command, y | ±0.5 m/s |
| Yaw rate command | ±1.0 rad/s |
| Commanded pelvis height | 0.18 – 0.73 m |

## V. RUNNING VLA POLICIES ON THE LOCOMOTION LAYER

The locomotion policy is exported to ONNX with a fixed interface of 71 inputs and 13 outputs and is loaded by a C++ controller that also holds the finite-state machine, the joint-level gains, and the operator interface. Metadata written into the exported file records the joint order, the PD gains, the action scales and offsets, the observation layout, and the commanded ranges the policy was trained on, together with a hash of that contract. A checking utility compares this metadata against the deployment configuration before a policy is run, which prevents a class of failure in which a policy is deployed against gains or a joint order it was not trained for.

Arm targets arrive on a dedicated topic that reuses the low-level command message type, so no new message definition is required. The controller treats the stream as live if a message has arrived within the last 200 ms. While it is live, the arm joint targets are taken from the message and slewed towards them at a rate limit of 20 rad/s, high enough to be transparent to the manipulation policy, and the proportional and derivative gains for the arm joints are taken from the same message, so the manipulation policy retains control of its own compliance. If the stream stops, the arms fall back to an operator-selected pose or to the resting carry pose and are moved there at a deliberately gentle rate.

Two manipulation policies were run through this interface without modifying the locomotion checkpoint between them. The first is GR00T N1.7 [5], [30] with a Cosmos-Reason2-2B backbone, fine-tuned on 605 demonstration episodes of fresh-produce handling collected on the same robot; the vision-language backbone was frozen, and only the action head was trained for ten thousand steps at a global batch size of 32. The second is π0.5 [26], adapted with low-rank adaptation on the same 605-episode dataset for twenty-two thousand steps. Both emit chunks of arm joint targets, and both were trained on a fixed base, with no knowledge of the locomotion policy. The manipulation performance of these models on this dataset is the subject of a companion study [31]; here they are used as interchangeable publishers of arm targets, and what is reported is the behavior of the locomotion layer beneath them. Because both policies were fine-tuned on the same demonstration dataset, the scope of the validation is limited accordingly: the interface is tested across model architectures and action heads, not across task domains.

## VI. EXPERIMENT

Simulation results use a fixed evaluation protocol: 1024 parallel environments on flat ground, twenty-second episodes, the upper-body pose curriculum pinned at full amplitude, and the same command distribution used at the end of training. Metrics follow those reported for the same class of controller in [2]: mean linear velocity error, mean yaw rate error, mean pelvis height error, symmetry loss, and living time, which saturates at the 20 s episode length when no falls occur. A fall rate expressed per environment-minute is also reported. Table IV reports the deployed policy under this protocol.

### *A. Deployed policy*

**TABLE IV:** Deployed policy under the evaluation protocol.

| Metric | Value |
|---|---|
| Linear velocity error (m/s) | 0.186 |
| Angular velocity error (rad/s) | 0.142 |
| Height error (m) | 0.016 |
| Living time (s) | 20.0 |
| Falls per environment-minute | 0.000 |

The deployed checkpoint was selected by evaluating every saved checkpoint from iteration 10,000 onward under this protocol and ranking by fall rate first and tracking error second, rather than by taking the final checkpoint. This mattered: the last checkpoint of the run was the weakest of those evaluated, and taking it would have shipped a policy with more than twice the standing base displacement of the one selected.

### *B. Ablations*

Each ablation changes exactly one design choice, holds everything else fixed, and trains for the same budget with the curriculum horizons compressed by a factor of two so that the reference variant still reaches the full task distribution within the shortened run. All variants are evaluated under the reference protocol, with only the trained policy varying; evaluating each under its own ablated configuration would change the test alongside the treatment. Because the ablations preserve the observation and action layout, the checkpoints are interchangeable at evaluation time. Table V reports the final numbers, and Fig. 4 reports the trajectory across training.

**TABLE V:** Matched-budget ablations, evaluated under the reference protocol. Lower is better except for living time.

| Variant | Lin. error (m/s) | Ang. err (rad/s) | Height error (m) | Sym. loss | Alive (s) | Falls/ env-min |
|---|---|---|---|---|---|---|
| Reference | 0.198 | 0.143 | 0.045 | 0.0001 | 20.0 | 0.00 |
| No symmetry | 0.221 | 0.178 | 0.063 | 115.4 | 20.0 | 0.00 |
| No height command | 0.328 | 0.281 | 0.100 | 0.0001 | 11.1 | 2.67 |
| No arm curriculum | 0.484 | 0.298 | 0.062 | 0.0001 | 7.4 | 6.06 |

Removing the upper-body pose curriculum is the most damaging of the three. A policy that has never been trained against arm motion fails to remain stable when evaluated under arm motion: living time falls from the full episode length to roughly seven seconds, the fall rate rises to six per environment minute, and velocity tracking error more than doubles. Removing the height command is also costly, and not only in height tracking. The variant loses close to half of its living time and falls at nearly three per environment minute. Part of that gap reflects the mismatch between a policy trained without height commands and a protocol that issues them, so the figure is best read as the cost of removing the command from the system rather than as an isolated measure of postural stability. The direction is nonetheless clear, and the magnitude places this component second only to the arm curriculum. Removing symmetry produces a different pattern. The symmetry loss rises by six orders of magnitude, as expected, while living time and fall rate are unchanged, and the tracking errors degrade by between 12% and 40%. Symmetry, on this robot and with this reward set, buys gait symmetry and an improvement in tracking and is not what keeps the robot upright.

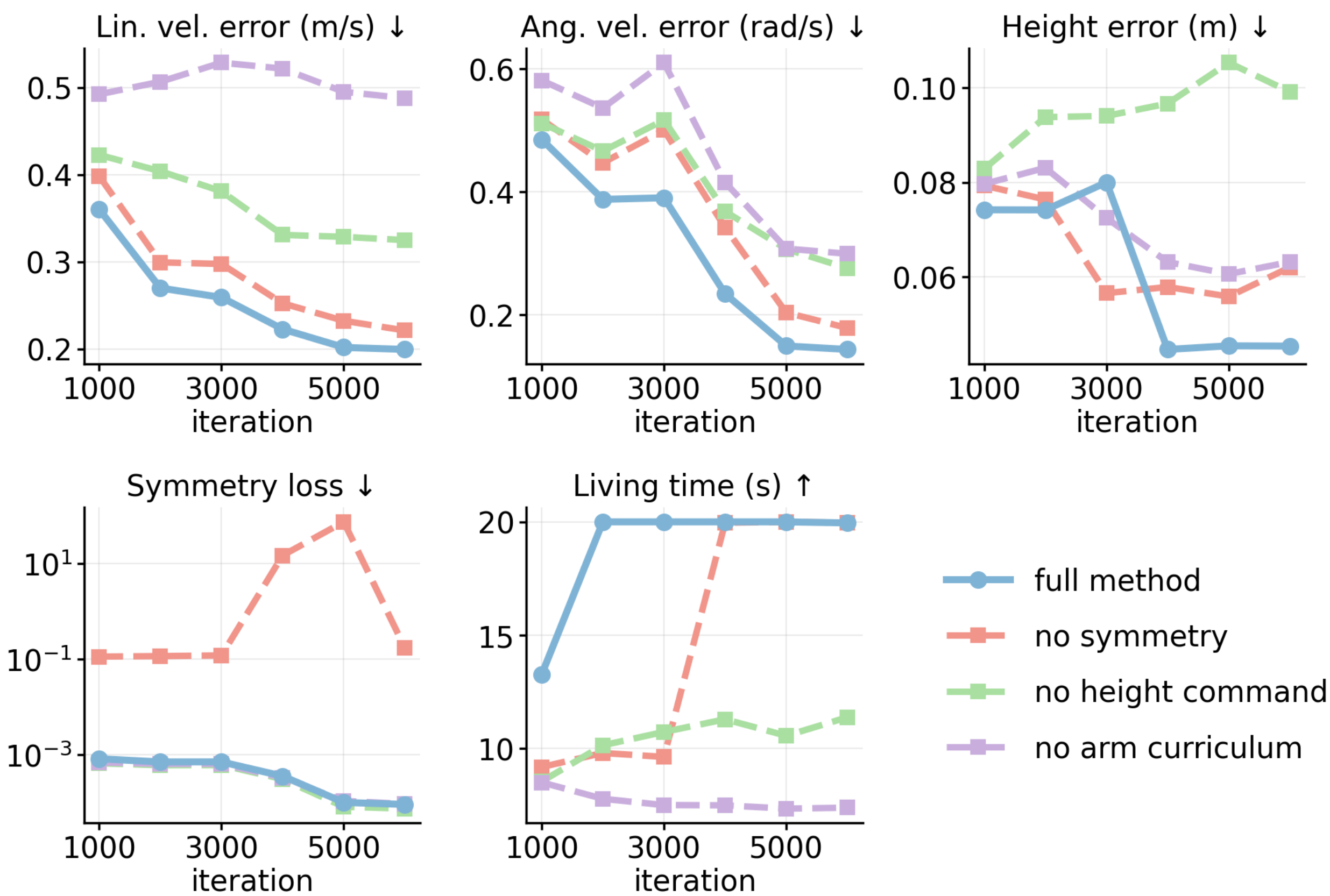

*Fig. 4. Ablations across training. One panel per metric, evaluated at saved checkpoints under the reference protocol. Each ablated variant removes one component and is plotted against the full method. Arrows in the panel titles give the direction of improvement. Symmetry loss uses a logarithmic scale.*

## *C. Operating region*

To map the conditions under which the deployed policy remains stable, the hand payload was swept against push magnitude under the same protocol, deliberately continuing past the ranges used in training so that the margin beyond the trained operating range is visible. Payload was held fixed per environment at values from zero to twice the training ceiling, and pushes were applied at magnitudes up to roughly twice those seen in training. Fig. 5 gives the resulting envelope.

Across the whole trained payload range, no falls occur at push magnitudes up to 1.0 m/s, which is above the 0.7 m/s used during training, and living time remains within half a second of the full episode. Falls appear only at the largest push and remain below 0.12 per environment minute inside the trained payload range. Beyond the training ceiling, the degradation is gradual rather than abrupt: at twice the ceiling, the policy still does not fall under pushes up to 0.7 m/s. Linear velocity tracking is almost flat across payloads and is marginally better in the middle of the trained payload distribution than at zero payload, which is consistent with the policy having been optimized over that distribution.

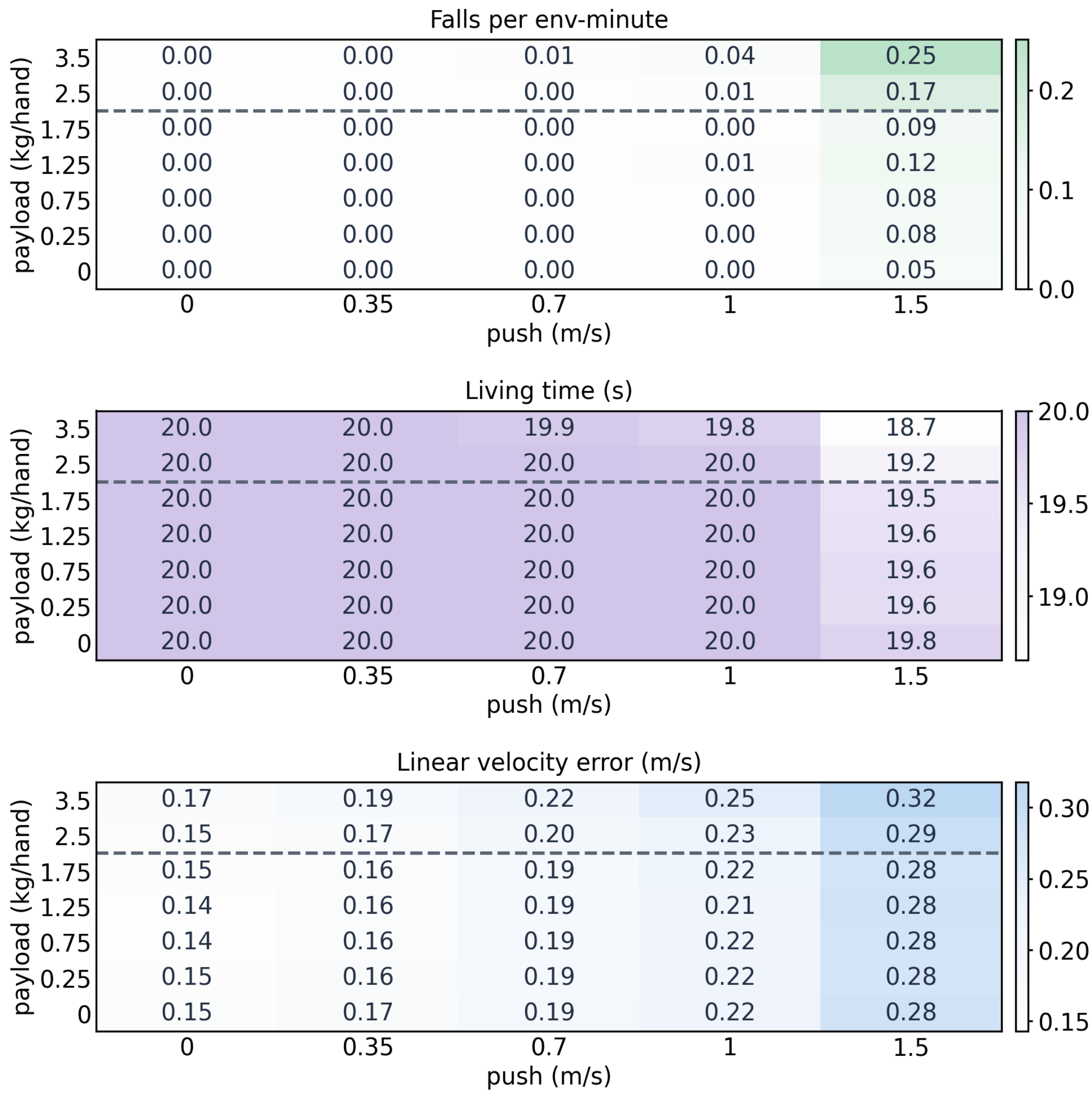

*Fig. 5. Operating region of the deployed policy, sweeping hand payload against push magnitude under the evaluation protocol. Top: falls per environment-minute. Middle: living time. Bottom: linear velocity error. The dashed line marks the payload ceiling used in training; rows above it lie outside the trained range and are included to show the margin.*

## D. Hardware

On hardware the policy was operated from a handheld remote, with forward speed limited at the controller to 0.5 m/s for manipulation work. The robot walked, turned, and held commanded crouches while carrying a loaded box in both hands and held the commanded height while the arms were driven through reaching poses. Both VLA models were run on the same locomotion checkpoint and completed reach, grasp, and place sequences on produce items without the locomotion policy being modified, restarted, or informed of the change. A safety gantry was attached throughout and remained slack; it carried no load during operation.

The box used in the carrying trials weighed 722.6 g loaded, comprising three 110 g printed cubes, three 40 g printed cubes, a 214 g wooden block, and the 58.55 g box itself, carried in two hands. Together with the 810 g hand assembly, half of the box places approximately 1.2 kg at each hand, within the payload range used in training. Fig. 6 shows the loaded box carried at three commanded heights, and Fig. 7 shows a grasp, transport, and place executed by a VLA while the locomotion policy holds the base.

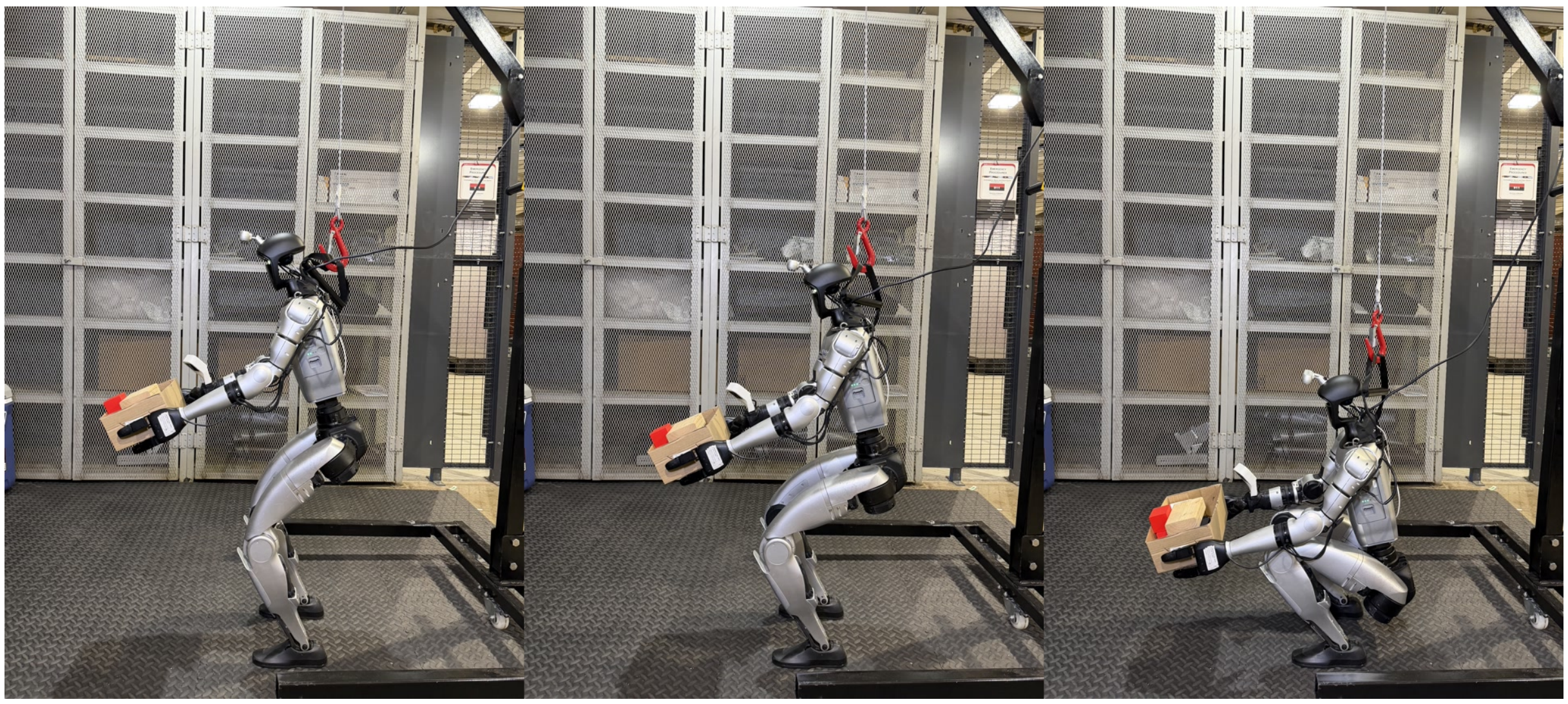

*Fig. 6. Carrying a 722.6 g loaded box in two hands at three commanded heights: full stand, half crouch, and full crouch. The contents and arm pose are identical across the three.*

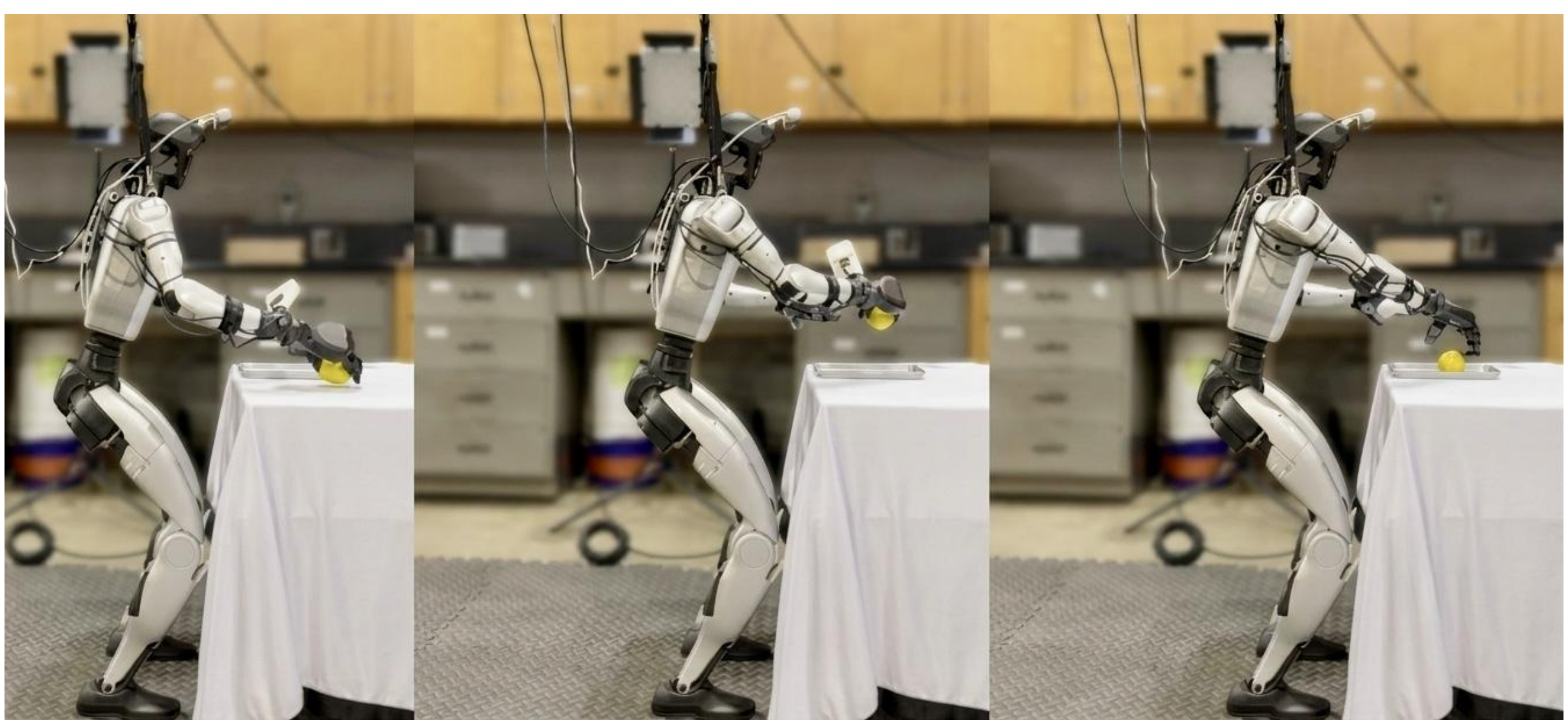

*Fig. 7. Grasp, transport, and place executed by a vision-language-action model while the locomotion policy holds the base. The locomotion policy receives no information about the manipulation task and was not modified between the two models tested.*

### *E. Cost of decoupling*

Decoupling is a design choice with a price, and that price was measured on the same robot. This study compared the deployed policy against a whole-body velocity policy trained on the same platform, which controls all twenty-three joints and therefore commands its own arms. To make the comparison fair, the commanded height of the decoupled policy was pinned at standing, and its arm disturbance was frozen so that both controllers performed the same task; the command distribution, pushes, friction, and hand payload were matched, with the payload event injected into the whole-body configuration, which had never been trained with one. Table VI gives the comparison.

**TABLE VI:** Decoupled versus whole-body control, matched conditions.

| Payload | Controller | Lin. error (m/s) | Ang. err (rad/s) | Tilt (deg) | Falls/ env-min |
|---|---|---|---|---|---|
| 0 kg | DUET (13 act.) | 0.214 | 0.155 | 9.65 | 0.00 |
| 0 kg | Whole-body (23 act.) | 0.163 | 0.160 | 1.66 | 0.00 |
| 1.75 kg | DUET (13 act.) | 0.188 | 0.151 | 9.72 | 0.00 |
| 1.75 kg | Whole-body (23 act.) | 0.178 | 0.176 | 1.73 | 0.00 |

Neither controller falls, including at a payload the whole-body policy never saw in training. The whole-body controller tracks straight-line velocity better, by around a quarter at zero payload, and it holds the torso far closer to vertical: a mean pelvis tilt of 1.7 degrees against 9.7 for the decoupled policy. That gap has a straightforward explanation. A whole-body controller can use the arms as angular-momentum actuators to damp torso motion, and a decoupled lower body cannot, because the arms are not available to it. The decoupled policy holds an advantage in yaw tracking once payload is applied: the whole-body policy degrades by about 10%, while the decoupled policy is unchanged, which is what the payload randomization was included to obtain. Anyone choosing a decoupled architecture in order to run an interchangeable manipulation policy should expect to pay for it in postural stability.

*F. Comparison with prior systems*

Table I sets out the properties on which the present system differs from prior loco-manipulation work. Table VII returns to those properties and records, for each, the measurement in this study that bears on it.

**TABLE VII:** Gaps identified in Table I and the supporting evidence in this study.

| Gap identified in Table I | Systems affected | How DUET addresses it | Evidence in this study |
|---|---|---|---|
| Arm targets reinterpreted by the learned controller | [3], [15], [18] | Targets copied into the outgoing message, only rate-limited | Section V: two VLAs on one checkpoint (VI-D) |
| Exchangeable manipulation policy not demonstrated | [2], [15], [19] | Any process publishing joint targets | GR00T N1.7 and π0.5, no retrain or restart (VI-D) |
| No commanded base height | [3], [16], [17] | Height command, 0.18–0.73 m | Height error 0.016 m (Table IV) |
| Arms driven by operator or IK solver | [2], [19] | Learned VLA | Section V; Fig. 7 |
| Robustness under payload not reported | [3], [15], [19] | Payload and push sweep past training range | No falls to 1.0 m/s across trained payload (Fig. 5) |
| Postural cost of decoupling unquantified | [2], [19] | Matched comparison on one platform | 9.7° vs 1.7° tilt (Table VI) |

## VII. DISCUSSION

Both manipulation policies drove the arms on a single locomotion checkpoint that was not modified, retrained, or restarted between them, and the lower body tracked commanded velocity, yaw rate, and pelvis height to within 0.186 m/s, 0.142 rad/s, and 0.016 m without falls. Earlier decoupled controllers also exclude the arms from the learned action space, but their arm targets come from an exoskeleton cockpit [2] or an inverse-kinematics solver [19], each a fixed part of the system. In DUET the source of those targets is unconstrained, because the locomotion policy is never informed of what produces them. Excluding the arms transfers the difficulty to training, and the ablations indicate where it lands. The upper-body pose curriculum accounts for most of it: without it, living time drops from 20.0 s to 7.4 s, and the fall rate reaches 6.06 per environment-minute. The height command is second: without it, living time is 11.1 s and the fall rate is 2.67 per environment-minute. Removing mirror symmetry affects neither measure and increases tracking error by 12–40%. HOMIE [2] also ablated these three components and reported similar living times across its curriculum and height-reward variants; in the present ablations, which remove each component outright, their contributions are unequal, and only the first is responsible for keeping the robot upright. Training against payload and hand inertia also explains the single metric on which the decoupled policy exceeds the whole-body baseline. At 1.75 kg per hand, the whole-body controller loses roughly 10% of its yaw tracking accuracy, and the decoupled policy loses none, consistent with the fact that only it was trained with hand payloads (Section VI-E). The cost of decoupling is postural. Mean pelvis tilt is 9.7° against 1.7°, and this gap follows from the architecture rather than from reward tuning, because the arms are not available to the decoupled policy as momentum actuators. Whole-body control is therefore preferable where torso stability is the binding constraint. Where the manipulation policy must be replaceable, and where that policy was trained on a fixed base and issues joint targets on the assumption that they are executed as written, DUET obtains that property at a cost of 8° in mean torso tilt.

## VIII. LIMITATIONS AND FUTURE DIRECTION

Several limitations bound these results. The torso stability gap is the most consequential: a mean pelvis tilt of 9.7° against 1.7° follows from the architecture rather than the reward set, since the arms are unavailable to the locomotion policy as momentum actuators. Its cost to manipulation is substantial. Over ten hardware trials per condition, GR00T N1.7 completed 8 of 10 grasps on a fixed base and 4 of 10 under the locomotion layer, and π0.5 completed 2 of 10 under the locomotion layer. The sample is small, and the two models are not separable at this size, but the drop relative to a fixed base is consistent across both. This study therefore establishes the locomotion layer and its interface, not end-to-end manipulation capability. Interchangeability also holds over a narrow range, as both policies were fine-tuned on the same 605-episode dataset, so the interface tolerates a model change rather than a task-domain change. All results come from a single seed; the curriculum and height-command effects are too large for seed variation to explain, but the 12–40% symmetry gap is not. Finally, the policy trains on flat ground, settles at a shallower crouch than the 0.18 m commanded floor, and models the hand as added mass rather than a full inertia tensor.

The main open problem is the drop in grasp success under a moving base, and the first step is to attribute it. Base tilt, standing drift, and the latency between a 15 Hz action chunk and a 50 Hz locomotion loop are separable causes, and logging pelvis pose against grasp outcome would identify which one dominates. Part of the gap may be recoverable without abandoning pass-through by allowing the locomotion policy to observe the published arm targets one cycle ahead while remaining forbidden to alter them, replacing disturbance rejection with anticipation. Fine-tuning the manipulation model on demonstrations collected under a moving base is the complementary route, since both models here were trained on a fixed base. Broader evidence for interchangeability requires drivers that differ in task domain and not only in architecture, including a non-learned publisher such as a teleoperation stream. Training across multiple seeds would place confidence intervals on the ablations, and extending the policy to uneven terrain requires either proprioceptive-only training or a second export contract for the perceptive variant.

## IX. CONCLUSION

This study introduced DUET, a decoupled loco-manipulation framework designed around a simple requirement: changing the manipulation policy should not require changing the controller that keeps the humanoid upright. DUET assigns lower-body stabilization and mobility to an RL locomotion policy while leaving the arm joints entirely under the control of an independent manipulation policy. Rather than learning coordination with a specific upper-body controller, the locomotion policy is trained to remain stable under the physical variability introduced by manipulation, including changes in arm configurations, payloads, inertial properties, and pelvis heights. Hardware experiments on a 23-DOF Unitree G1 show that this separation holds in practice, with a single locomotion checkpoint operating beneath two independently trained vision-language-action models without locomotion retraining, policy-specific adaptation, or changes to the execution interface. Ablation studies show that exposure to diverse upper-body poses is critical to achieving this robustness, while pelvis-height conditioning and mirror symmetry provide additional benefits for stability and tracking. Push and payload experiments further establish the learned controller's operating range, and the matched whole-body comparison reveals the principal trade-off of explicit decoupling: reduced postural coordination in exchange for greater modularity. The key result is therefore not simply that locomotion and manipulation can run concurrently, but that the manipulation policy can change while the learned locomotion layer remains fixed. DUET achieves this by training the lower body against the physical consequences of manipulation rather than against the internal behavior of a particular manipulation model, allowing locomotion to function as a reusable stabilization layer beneath independently developed manipulation policies. As vision-language-action and other foundation models continue to evolve, this separation provides a practical path for integrating new manipulation capabilities into mobile humanoids without repeatedly redesigning or retraining the locomotion system, shifting humanoid loco-manipulation toward a more modular architecture in which manipulation models can evolve independently while a common locomotion layer continues to provide mobility and balance.

**CRediT authorship contribution statement**

**Kapalik Khanal:** Conceptualization, Methodology, Software, Validation, Formal analysis, Data curation, Visualization, Writing - original draft, Writing - review & editing. **Biswash Khatiwada:** Conceptualization, Data curation, Writing - original draft, Writing - review & editing. **Stephen Afrifa:** Conceptualization, Data curation, Writing - review & editing. **Fan Wu:** Writing - review & editing. **Ramesh Bahadur Bist:** Conceptualization, Methodology, Data curation, Supervision, Project administration, Funding acquisition, Writing - original draft, Writing - review & editing.

**Acknowledgments**

This project was supported by North Carolina State University's College of Agriculture and Life Sciences and the Department of Biological and Agricultural Engineering Faculty startup fund.

**Data availability**

The training configuration, export and verification tooling, and deployment controller are available at https://github.com/bae-air-lab/DUET and video can be found here: https://www.youtube.com/watch?v=pHMaEnotOqI.

**Declaration of competing interests**

The authors declare that they have no known competing financial interests or personal relationships that could have appeared to influence the work reported in this paper.

**Declaration of Artificial Intelligence use and integration.**

Grammarly was used solely for language editing to improve the manuscript's clarity, grammar, and consistency.

**Conflict of Interest**

The authors declare no conflict of interest.

**REFERENCES**

[1] C. Lu, X. Cheng, J. Li, S. Yang, M. Ji, C. Yuan, G. Yang, S. Yi, and X. Wang, "Mobile-TeleVision: predictive motion priors for humanoid whole-body control," in Proc. IEEE ICRA, 2025. arXiv:2412.07773.

[2] Q. Ben, F. Jia, J. Zeng, J. Dong, D. Lin, and J. Pang, "HOMIE: humanoid loco-manipulation with isomorphic exoskeleton cockpit," in Proc. RSS, 2025. arXiv:2502.13013.

[3] H. Ha, Y. Gao, Z. Fu, J. Tan, and S. Song, "UMI on Legs: making manipulation policies mobile with manipulation-centric whole-body controllers," in Proc. CoRL, 2024. arXiv:2407.10353.

[4] Z. Fu, T. Z. Zhao, and C. Finn, "Mobile ALOHA: learning bimanual mobile manipulation with low-cost whole-body teleoperation," in Proc. Conf. Robot Learning (CoRL), 2024. arXiv:2401.02117.

[5] NVIDIA, J. Bjorck et al., "GR00T N1: an open foundation model for generalist humanoid robots," arXiv:2503.14734, 2025.

[6] C. C. Kemp, A. Edsinger, and E. Torres-Jara, "Challenges for robot manipulation in human environments," IEEE Robot. Autom. Mag., vol. 14, no. 1, pp. 20–29, 2007.

[7] N. Rudin, D. Hoeller, P. Reist, and M. Hutter, "Learning to walk in minutes using massively parallel deep reinforcement learning," in Proc. CoRL, 2022, pp. 91–100.

[8] V. Makoviychuk et al., “Isaac Gym: high performance GPU-based physics simulation for robot learning,” in Proc. NeurIPS Datasets and Benchmarks Track, 2021.
[9] J. Schulman, F. Wolski, P. Dhariwal, A. Radford, and O. Klimov, “Proximal policy optimization algorithms,” arXiv:1707.06347, 2017.
[10] J. Hwangbo et al., “Learning agile and dynamic motor skills for legged robots,” Sci. Robot., vol. 4, no. 26, eaau5872, 2019.
[11] J. Lee, J. Hwangbo, L. Wellhausen, V. Koltun, and M. Hutter, “Learning quadrupedal locomotion over challenging terrain,” Sci. Robot., vol. 5, no. 47, eabc5986, 2020.
[12] I. Radosavovic, T. Xiao, B. Zhang, T. Darrell, J. Malik, and K. Sreenath, “Real-world humanoid locomotion with reinforcement learning,” Sci. Robot., vol. 9, no. 89, eadi9579, 2024.
[13] K. Black et al., “π0: a vision-language-action flow model for general robot control,” arXiv:2410.24164, 2024.
[14] M. J. Kim et al., “OpenVLA: an open-source vision-language-action model,” in Proc. CoRL, 2024.
[15] X. Cheng, Y. Ji, J. Chen, R. Yang, G. Yang, and X. Wang, “Expressive whole-body control for humanoid robots,” in Proc. RSS, 2024.
[16] Z. Fu, Q. Zhao, Q. Wu, G. Wetzstein, and C. Finn, “HumanPlus: humanoid shadowing and imitation from humans,” in Proc. CoRL, 2024.
[17] T. He et al., “OmniH2O: universal and dexterous human-to-humanoid whole-body teleoperation and learning,” in Proc. CoRL, 2024.
[18] T. He et al., “HOVER: versatile neural whole-body controller for humanoid robots,” in Proc. IEEE ICRA, 2025.
[19] NVIDIA, "GR00T Whole-Body Control," GitHub repository, 2025. [Online]. Available: https://github.com/NVlabs/GR00T-WholeBodyControl (accessed Jan. 2026).
[20] J. Tobin, R. Fong, A. Ray, J. Schneider, W. Zaremba, and P. Abbeel, “Domain randomization for transferring deep neural networks from simulation to the real world,” in Proc. IEEE/RSJ IROS, 2017, pp. 23–30.
[21] A. Kumar, Z. Fu, D. Pathak, and J. Malik, “RMA: rapid motor adaptation for legged robots,” in Proc. RSS, 2021.
[22] E. Todorov, T. Erez, and Y. Tassa, “MuJoCo: a physics engine for model-based control,” in Proc. IEEE/RSJ IROS, 2012, pp. 5026–5033.
[23] K. Zakka, Q. Liao, B. Yi, L. Le Lay, K. Sreenath, and P. Abbeel, "mjlab: a lightweight framework for GPU-accelerated robot learning," arXiv:2601.22074, 2026.
[24] K. Zakka et al., “MuJoCo Playground,” arXiv:2502.08844, 2025.
[25] A. Brohan et al., “RT-2: vision-language-action models transfer web knowledge to robotic control,” in Proc. CoRL, 2023. arXiv:2307.15818.
[26] Physical Intelligence, K. Black et al., “π0.5: a vision-language-action model with open-world generalization,” arXiv:2504.16054, 2025.
[27] C. Chi et al., “Diffusion policy: visuomotor policy learning via action diffusion,” in Proc. RSS, 2023.
[28] T. Z. Zhao, V. Kumar, S. Levine, and C. Finn, “Learning fine-grained bimanual manipulation with low-cost hardware,” in Proc. RSS, 2023.
[29] Unitree Robotics, “Unitree G1 humanoid robot,” 2024. [Online]. Available: https://www.unitree.com/g1

[30] NVIDIA, "GR00T-N1.7-3B," Hugging Face model card, 2026. [Online]. Available: https://huggingface.co/nvidia/GR00T-N1.7-3B (accessed April 2026).
[31] K. Khanal, B. Khatiwada, S. Afrifa, S. Exum, F. Wu, and R. B. Bist, "Vision-language-action modeling for humanoid robotic handling of fresh agricultural produce," 2026. doi:10.2139/ssrn.7175038.